\documentclass[preprint,review,12pt]{elsarticle}

\usepackage{epsfig}

\usepackage{amssymb}
\usepackage{amsmath,amsfonts}
\usepackage{algorithmic}
\usepackage{array}
\usepackage[caption=false,font=normalsize,labelfont=sf,textfont=sf]{subfig}
\usepackage{textcomp}
\usepackage{stfloats}
\usepackage{url}
\usepackage{verbatim}
\usepackage{graphicx}
\usepackage{amsmath}
\usepackage{amssymb}
\usepackage{xcolor}
\usepackage{hyperref}
\usepackage{svg}
\usepackage{calc}
\usepackage{multicol}
\usepackage{accents}
\usepackage{footnote}
\usepackage{multirow}
\usepackage{amstext}
\usepackage{amsfonts}
\usepackage{stmaryrd}
\usepackage{textcomp}
\usepackage[mathscr]{euscript}
\usepackage{bigints}
\usepackage{upgreek}
\usepackage[percent]{overpic}
\usepackage{soulutf8}
\usepackage{bm}
\usepackage{physics}
\usepackage{mathtools}
\usepackage{caption}
\usepackage{yhmath}
\usepackage{tikz}
\usepackage{comment}
\usepackage{cellspace}
\usepackage{hyperref}
\usepackage{tabularx}
\usepackage{adjustbox}
\usepackage{colortbl}
\usepackage{mwe}
\usepackage{xcolor}
\usepackage{tabularx}
\usepackage{multirow} % For multi-row cells
\usepackage{soul}
\usepackage{tabularx}
\usepackage{booktabs} % For better table formatting
\usepackage{graphicx}
\usepackage{booktabs}
\usepackage{makecell}
\usepackage{colortbl}
\usepackage{multirow}
\usepackage{xcolor}
\usepackage{pifont}
\usepackage{bm}
\usepackage{mathrsfs} % For an alternative calligraphic style
\usepackage{soul}
\usepackage{amssymb}

\usepackage[capitalise,noabbrev]{cleveref}
\crefname{para}{Paragraph}{Paragraphs}
\crefname{chapter}{Chapter}{Chapters}
\crefname{section}{Section}{Sections}
\crefname{subsection}{Section}{Sections}
\crefname{subsubsection}{Section}{Sections}
\crefname{equation}{Equation}{Equations}
\crefname{figure}{Figure}{Figures}
\crefname{table}{Table}{Tables}

\setcitestyle{numbers,square}
\journal{Pattern Recognition}

\begin{document}

\begin{frontmatter}

\title{iMatcher: Improve matching in point cloud registration via local-to-global geometric consistency learning}
% Fully Differentiable Assignment Matrix with Geometric Constraints (iMatcher )

\author[Karim Slimani]{Karim Slimani}
\ead{{karim.slimani@isir.upmc.fr}}
\author[Karim Slimani]{Catherine Achard}
\author[Karim Slimani]{Brahim Tamadazte}

\affiliation[Karim Slimani]{organization={ISIR, Sorbonne Université, CNRS UMR 7222, INSERM U1150},%Department and Organization
            addressline={4 place Jussieu}, 
            city={Paris},
            postcode={75005}, 
            % state={},
            country={France}}

%% Abstract
\begin{abstract}
This paper presents \textit{iMatcher}, a fully differentiable framework for feature matching in point cloud registration. The proposed method leverages learned features to predict a geometrically consistent confidence matrix, incorporating both local and global consistency. First, a local graph embedding module leads to an initialization of the score matrix. A subsequent repositioning step refines this matrix by considering bilateral source-to-target and target-to-source matching via nearest neighbor search in 3D space. The paired point features are then stacked together to be refined through global geometric consistency learning to predict a point-wise matching probability. Extensive experiments on real-world outdoor (KITTI, KITTI-360) and indoor (3DMatch) datasets, as well as on 6-DoF pose estimation (TUD-L) and partial-to-partial matching (MVP-RG), demonstrate that iMatcher significantly improves rigid registration performance. The method achieves state-of-the-art inlier ratios, scoring 95\%–97\% on KITTI, 94\%–97\% on KITTI-360, and up to 81.1\% on 3DMatch, highlighting its robustness across diverse settings. %\textcolor{blue}{The source code will be publicly available after publication.}
\end{abstract}

% %%Graphical abstract
% \begin{graphicalabstract}
% \begin{center}
    
% \centerline{\includegraphics[width=1\columnwidth]{figures/pipeline.pdf}}

% \end{center}
% \end{graphicalabstract}

% %% Research highlights
% \begin{highlights}
% \item A fully differentiable framework for feature matching in point cloud registration.
% \item Local and global geometric consensus learning, enabling structured and interpretable prediction of the soft assignment matrix.
% \item State-of-the-art inlier ratio demonstrated through experimental validation on diverse point cloud registration datasets, including scene-centric (KITTI, KITTI-360, and 3DMatch) and object-centric (TUD-L and MVP-RG) benchmarks.
% \item Versatility and consistency shown through deployment within various point cloud registration architectures

% \end{highlights}


\end{frontmatter}
% -------------------
\section{Introduction and related work}
% -------------------
%
\begin{figure}[!t]
\centerline{\includegraphics[width=.95\columnwidth]{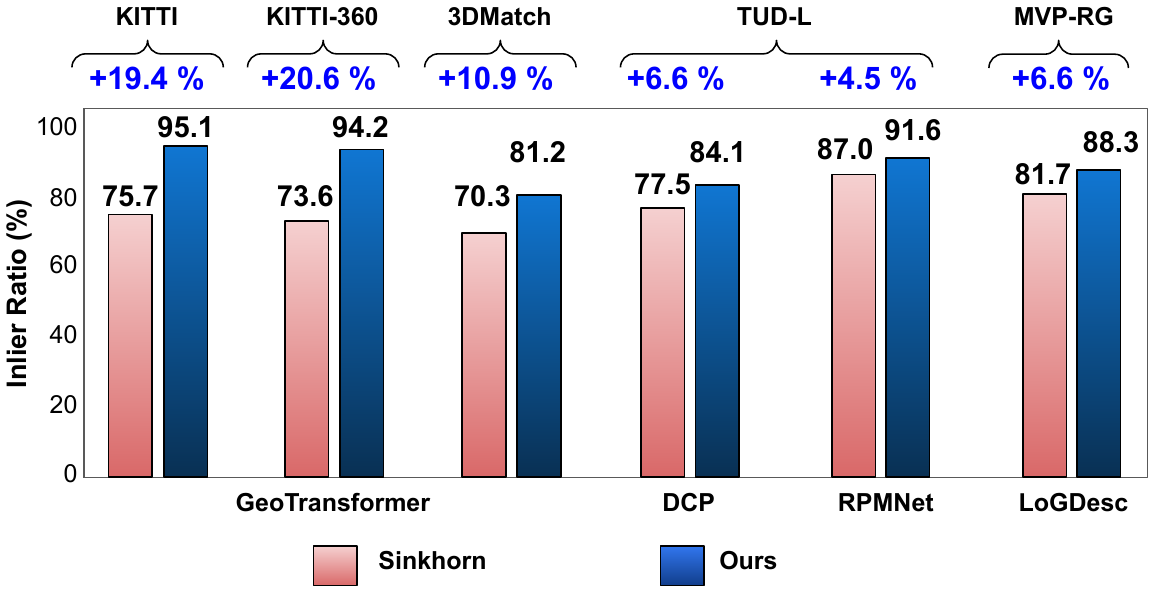}
}
\caption{\textit{iMatcher} impacts various SOTA methods for Inlier Ratio across KITTI, KITTI-360, 3DMatch, TUD-L, and MVP-RG datasets.}
\label{fig.abstract}
\end{figure}
\label{sec:intro}

Tridimensional point cloud registration (PCR) is a critical problem in robotics and computer vision with multiple applications in autonomous driving~\cite{shi2023rdmnet}, object grasping and manipulation~\cite{jiang2024se}, 3D reconstruction~\cite{pointdt}, and inter-modal medical alignment~\cite{liu2022fusion,choi2025deep} and many other domains. A PCR problem aims to estimate the rigid transformation matrix that aligns a source point cloud with a target one. Such an operation presents multiple challenges due to the inherent disorder of point clouds, potential occlusions during real-world data acquisition, and sensor-induced noise, among other factors. Consequently, estimating a precise transformation requires extracting accurate correspondences while maximizing the inlier ratio. The related literature~\cite{lyu2024rigid} identifies two main categories: optimization-based techniques and learning-based pipelines. \\ 
\textbf{Optimization-based techniques:} Before the emergence of deep learning-based methods, PCR approaches relied on different optimization-based techniques.  Iterative Closest Point (ICP)~\cite{besl1992method} was the standard PCR solution for many years. It operates in two different steps: first, the closest correspondences are searched in the Euclidean space, and then the rigid transformation is computed iteratively until convergence. Some other derivatives have been developed based on ICP to address its dependency on initialization.  For instance, Go-ICP~\cite{yang2013go} exploits the geometry
of $SE(3)$ using Branch-and-Bound (BnB) to guarantee optimality regardless of the initialization quality. The authors of~\cite{zhang2021fast} propose to improve ICP’s robustness by introducing a new metric error based on Welsch’s function~\cite{holland1977robust} while enhancing its efficiency through the Anderson-accelerated majorization-minimization strategy~\cite{walker2011anderson}. In~\cite{lv2023kss}, the authors introduced a new ICP version (KSS-ICP) relying on a Kendall Shape Space (KSS)-based representation of the point clouds to limit the impact of density gradients, translation, scale, and rotation. Nevertheless, deep learning models have emerged as the dominant paradigm for PCR problems, with current research primarily centered on advancing these models, as detailed in the following.\\
\textbf{Deep learning-based techniques:} A first category of deep learning-based PCR methods is trained to explicitly predict the rigid transformation matrix without requiring a matching process~\cite{aoki2019pointnetlk,li2021pointnetlk}. For instance, PointNetLK~\cite{aoki2019pointnetlk} extracts global features using PointNet~\cite{qi2016pointnet} and proposes a modified Lucas-Kanade algorithm incorporated into the PointNet baseline with a recurrent neural network to predict a rigid transformation. This is achieved via a stochastic gradient method in the computation of the Jacobian of the global feature vector. The inverse of the Jacobian matrix is then used to incrementally refine the pose by recomputing the optimal twist parameters at each iteration. However, the method suffers from numerical instabilities. In~\cite{li2021pointnetlk}, the authors employed an analytical Jacobian to address this issue, enhancing the method's generalization capability. Conversely, correspondence-based methods learn to predict soft assignment matrices for pairing two-point clouds~\cite{geotransformer,slimani2024rocnet++,shi2023rdmnet}. The rigid transformation is then estimated using the learned putative correspondences, mostly via RANSAC. For instance, GeoTranformer~\cite{geotransformer} proposed an alternative to improve efficiency using a Local-to-Global (LGR) scheme. At the same time, RoCNet++~\cite{slimani2024rocnet++} introduced a Farthest Sampling-guided Registration (FSR) technique to provide precise alignment in a few iterations. A common aspect of both correspondence-free and correspondence-based architectures is the use of transformers~\cite{vaswani2017attention} to enhance local feature representation~\cite{dcp,geotransformer,liu2023regformer,parenet,slimani2024rocnet++}. For point clouds with \( N \) points, transformers incur a computational complexity of \( O(N^2) \), making them naturally unsuitable for large-scale data~\cite{kitti,kitti360,zeng20173dmatch}. Cofinet and GeoTransformer~\cite{yu2021cofinet,geotransformer} follow a coarse-to-fine strategy to overcome this limitation. Once the features of the input point clouds are extracted, the raw points are downsampled to produce coarser-scale superpoints, which are then fed into the transformer. The aggregated features from the transformer serve to establish superpoint-level correspondences. Local patches are subsequently formed by selecting the nearest neighbors of each matched superpoint to find dense point correspondences within the resulting subsamples.

Recently, diffusion models have emerged as a credible solution for PCR problems. The diffusion process can either be applied directly to the transformation or to the score matrices. For example, the authors of~\cite{jiang2024se} achieved significant improvements on learning baselines (i.e., DCP~\cite{dcp} and RPMNet~\cite{yew2020rpm}) for 6DOF pose estimation by incorporating a forward and backward diffusion process on the transformation learning, exploiting the \emph{Lie Algebra} associated with the $SE(3)$ space. Similarly, Diff-Reg~\cite{wu2024diff} proposes to estimate point matching via a denoising diffusion model on the doubly stochastic matrix space.

This paper presents a novel learning framework for partially overlapping point cloud matching. The proposed method, called \textit{iMatcher}, aims to estimate a doubly stochastic mapping function by exploiting both the local and global structures of each point cloud. To achieve this, \textit{iMatcher} generates graphs within the descriptor space and enhances features using convolutional networks to estimate an initial correspondence matrix. A weighted Singular Value Decomposition (SVD) is then used to determine a pre-alignment transformation, enabling a repositioning step that warps the source point cloud in the target space. Thus, a bilateral matching process is performed to incorporate global spatial consistency and assess the probability of each source and target point being an inlier. Finally, the correspondence matrix is refined using these matchability scores to obtain the final score matrix. The proposed \textit{iMatcher} method is evaluated on multiple datasets and integrated into various frameworks. Experiments on real-world outdoor datasets (KITTI, KITTI-360), indoor datasets (3DMatch), 6-DoF pose estimation (TUD-L), and partial-to-partial matching (MVP-RG) show that \textit{iMatcher} significantly improves rigid registration performance. As shown in Figure~\ref{fig.abstract}, \textit{iMatcher} achieves state-of-the-art inlier ratios, ranging from $95\%$ to $97\%$ on KITTI, $94\%$ to $97\%$ on KITTI-360, and up to $81.1\%$ on 3DMatch, highlighting its effectiveness across diverse scenarios.
% --------------
\section{Problem formulation}
% --------------
%
Given two overlapping point clouds $\boldsymbol{X}  \in\mathbb{R}^{M \times 3}$ and $\boldsymbol{Y} \in\mathbb{R}^{N \times 3} $, PCR aims to estimate the homogeneous $4\times4$ transformation matrix $\mathbf{T}$ that aligns at best $\boldsymbol{X}$ and $\boldsymbol{Y}$. The matrix $\mathbf{T}$ is defined as follows:
\begin{equation}
\mathbf{T}_{4\times4} =
\begin{bmatrix}
   \mathbf{R}_{3\times3} & \mathbf{t}_{3\times1} \\
   \mathbf{0}_{1\times3}& 1 \\
\end{bmatrix}
\end{equation}
where \( \mathbf{R} \in SO(3) \) and \( \mathbf{t} \in \mathbb{R}^3 \) are the desired rotation matrix and translation vector respectively. This step is preceded by the point-matching task, which estimates a mapping function $\phi: \boldsymbol{X} \to \boldsymbol{Y} \cup \{\varnothing\}$ used to compute the transformation.
Ideally, this function must verify:
\begin{equation}
\phi(\boldsymbol{x}_i) =
\begin{cases}
\arg\min\limits_{\boldsymbol{y}_j \in \boldsymbol{Y}} \| \mathbf{R} \boldsymbol{x}_i + \mathbf{t}  - \boldsymbol{y}_j \|_2, & \text{if } \| \mathbf{R} \boldsymbol{x}_i + \mathbf{t}  - \boldsymbol{y}_j \|_2 < \beta, \\
\varnothing, & \text{otherwise}.
\end{cases}
\end{equation}
where \( \beta \) is a threshold to filter points without correspondences. In practice, PCR models are trained to predict a soft matrix ${\boldsymbol{S}}$,
where each score ${\boldsymbol{S}}_{ij}$ represents the probability of pairing point $\
\boldsymbol{x}_i \in \boldsymbol{X}$ with point $\boldsymbol{y}_j \in \boldsymbol{Y}$.
\section{Motivation}\label{imatcher_motivation}
Despite the widespread success of the Sinkhorn algorithm in point cloud matching, its reliance on iterative optimization over an entropic regularization objective often limits interpretability and imposes computational overhead. In contrast, we propose a novel alternative that leverages both local and global geometric coherence as a foundational principle for match scoring. Our approach, instantiated in the \textit{iMatcher} framework, departs from traditional optimization by directly constructing a score matrix informed by the spatial structure of the point clouds.

This design enables several compelling advantages. Notably, our method supports progressive feature refinement during matching, laying the groundwork for future integration with score-based diffusion models. Such models, which iteratively denoise and evolve features across noise steps, align naturally with our framework’s architecture and do not require additional mechanisms, such as Transformers~\cite{vaswani2017attention}, to facilitate feature evolution.

By embedding geometric consistency into both the local and global stages of the matching process, iMatcher predicts a confidence matrix as a relaxed alternative to doubly stochastic constraints, obtained by modulating the \textit{dual-softmax} of the initial score matrix with a matchability matrix whose values lie between 0 and 1. The method begins by initializing a score matrix using a local graph embedding module, which enhances the feature representations of both source and target point clouds. These enriched features are then projected via an inner product to produce an initial estimate of the score matrix. A repositioning step is then carried out using a weighted Procrustes alignment, guided by the initial score matrix. This enables bilateral source-to-target and target-to-source matching through 3D nearest neighbor search. The resulting paired point features are stacked and processed to learn global geometric consistency, ultimately providing a point-wise matching probability matrix. In the final step, iMatcher fuses the initial score matrix with the matchability matrix to predict the final soft assignment matrix.
%
%--------
\section{Proposed Method}\label{sec:method}
\begin{figure}[!h]
\centerline{\includegraphics[width=1.\columnwidth]{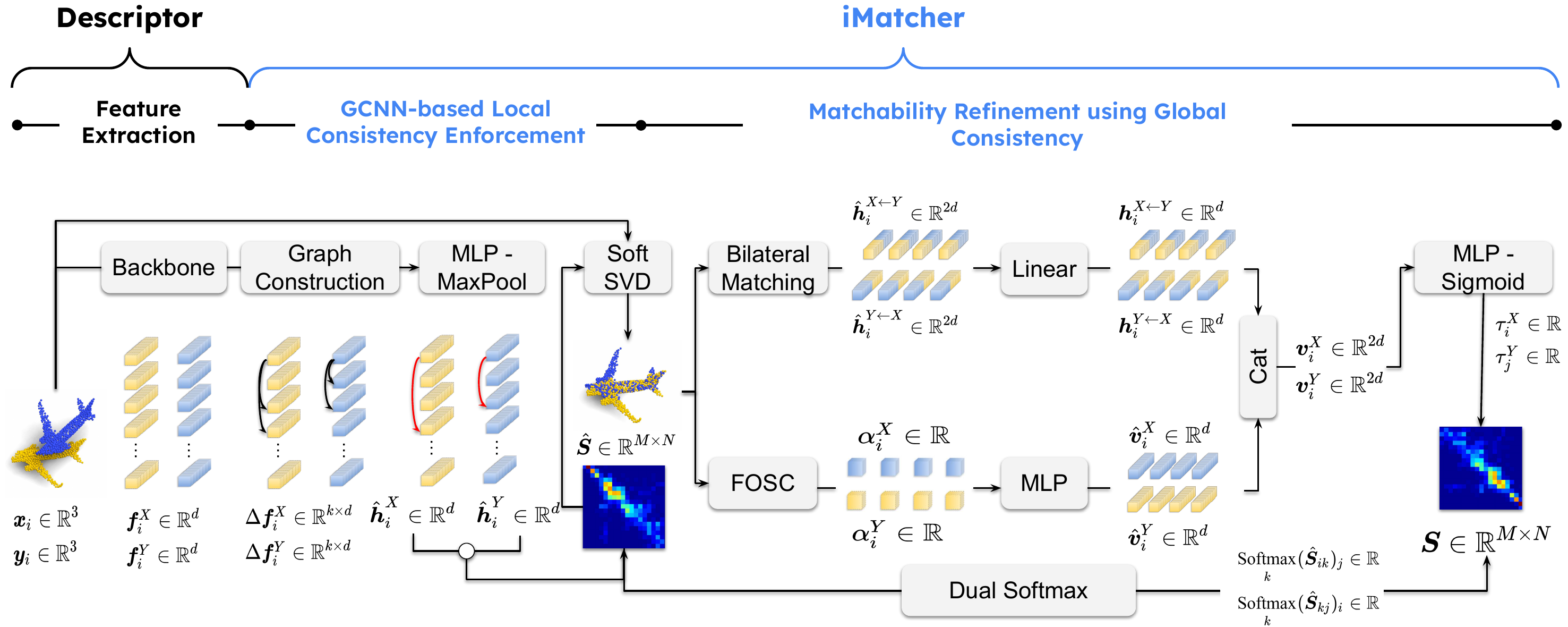}}
\caption{Overall \textit{iMatcher} pipeline. GCNN and FOSC stand for Graph CNN and First-Order Spatial Compatibility, respectively. %The MLP modules are $\bm{h}^1_\theta$, $\bm{h}^2_\theta$  and $\bm{h}^3_\theta$ in that order. 
}
\label{imatcher_fig_architecture}
\end{figure}

Let us consider $\bm{X} \in \mathbb{R}^{M\times3}$, $\bm{Y} \in \mathbb{R}^{N\times3}$, and their associated features  $\bm{F}^X \in \mathbb{R}^{M\times D}$, $\bm{F}^Y  \in \mathbb{R}^{N\times D}$ respectively. The proposed \textit{iMatcher} module aims to predict a score matrix ${\bm{S}}\in \mathbb{R}^{M\times N}$ that maps each inlier source point to its nearest neighbor in the target point cloud under the ground-truth transformation. 
\textit{iMatcher} consists of three key components: \textbf{a graph CNN-based local consistency enforcement}, \textbf{a correspondence refinement using global consistency} and \textbf{soft assignment matrix using local and global information fusion}. The first component leverages local graph convolutions to incorporate information from the immediate neighborhood of each point when computing matching scores. The second component estimates a confidence score for each source and target point based on global consistency. The last module refines the score matrix by fusing information from the two previous blocks, ensuring robust performance even in the presence of noise and significant occlusions. The overall pipeline is depicted in~\cref{imatcher_fig_architecture}, and detailed components are described in~\cref{imatcher_subsub_gcnn,imatcher_subsub_matchabili,imatcher_final_matrix} 
%
% --------
\subsection{Graph CNN-based local consistency enforcement}\label{imatcher_subsub_gcnn}
To capture local topological relationships, we construct a local graph using the \textit{EdgeConv} operation proposed by~\cite{dgcnn}, and collect, for each point $\bm{x}_i$, a subsample $\bm{P}^X_i \in \mathbb{R}^{k\times d} $ composed of the $K$ nearest neighbors features of $\bm{f}^X_i$ in the descriptor space. A simple concatenation of $\bm{f}^X_i$ and its relative coordinate-wise displacement in the feature space with all $\bm{f}^X_j \in \bm{P}^X_i$ allows for enhancing the structure of the local features. The resulting tensors are defined as follows:
\begin{equation}\label{imatcher_eq_intermediate_matrix}
 \Delta \bm{f}^X_i = \Big[ \delta f^X_{i1},...\delta f^X_{ij},...,\delta f^X_{ik}  \Big] ~, \Delta \bm{f}^X_i  \in \mathbb{R}^{k\times 2d}
\end{equation}
where $\delta f^X_{ij} \in \mathbb{R}^{2d}$ is defined as: 
\begin{equation}\label{imatcher_eq_intermediate_matrix}
\delta f^X_{ij} \coloneqq \left[ f^X_i, f^X_i - f^X_j \right] \in \mathbb{R}^{2d}
\end{equation}
Let $\hat{\bm{v}}^X_i \in \mathbb{R}^d$ be an updated feature vector for the $i$-th point in the set $X$, and let ${\bm{h}}^{X \leftarrow Y}_i \in \mathbb{R}^d$ be the projected feature incorporating information from the corresponding point in $Y$. Define the concatenated feature vector:
\[
\mathbf{H}^X_i = \bm{h}^1_\theta(\Delta \bm{f}^X_i), \quad \mathbf{H}^X_i \in \mathbb{R}^{k \times 2d}
\] 
where $\bm{h}^1_\theta$  is a three-layer MLP, with fully connected layers of output dimensions \(d\), \(d\), and \(2d\), respectively, each followed by \textit{Group Normalization} and a \textit{ReLU} activation function. The output features matrix  $\mathbf{H}^X_i$ is then pooled over the first dimension with a \textit{MaxPool} operation to get the local structure descriptors $\bm{h}^X_i \in \mathbb{R}^{2d}$. A linear projection is then learned to recover the original dimension of the features $\hat{\bm{h}}^X_i \in \mathbb{R}^{d}$ to be used for a point-wise matchability score estimation as explained in~\cref{imatcher_subsub_matchabili}. The same procedure is applied to $\bm{Y}$ to get $\hat{\bm{h}}^ Y _j \in \mathbb{R}^{d}$. We also compute an intermediate assignment score matrix $\hat{\bm{S}} \in \mathbb{R}^{M\times N}$, where: 
\begin{equation}\label{imatcher_eq_intermediate_matrix}
\hat{\bm{S}}_{ij} =  (\hat{\bm{h}}^X_i)^T   \hat{\bm{h}}^Y_j  ~~; ~~ \forall (i,j) \in M\times N
\end{equation}
%
% --------------
\subsection{Matchability refinement using global consistency}
This module draws inspiration from the 2D image feature matching approach LightGlue~\cite{lindenberger2023lightglue}, which estimates a matchability score for each point of $\bm{X}$ and $\bm{Y}$ by applying a linear layer to the input features. Notably, the matchability score for points in $\bm{X}$ (and similarly for $\bm{Y}$) is calculated using information exclusively from other points within $\bm{X}$ (or $\bm{Y}$). In contrast, \textit{iMatcher} considers $\bm{X}$ and $\bm{Y}$ points to compute the matchability score of points in $\bm{X}$ (respectively $\bm{Y}$). This allows both feature similarity and spatial geometry constraints to be used in estimating these scores.\\

%
% ------------
\textbf{Correspondences repositioning using differentiable SVD} \label{imatcher_subsub_repos}
To predict the probability of each source point $\bm{x}_i$ being an inlier (\textit{i.e.}, having a corresponding match in the target), incorporating the most similar features among the points of $\bm{Y}$ can significantly enhance the estimation. A straightforward approach would be to rely on $\hat{\bm{S}}$ and perform a \textit{top-1} score search to establish correspondences. However, instead of directly using $\hat{\bm{S}}$  for correspondence selection, we propose using it for a pre-alignment step. Specifically, we utilize a differentiable weighted SVD~\cite{dcp} to estimate a rigid transformation that warps the source point cloud $\bm{X} = \{ \bm{x}_i \}_{i=1}^N$ into the target space. Each point $\bm{x}_i$ is transformed into $\hat{\bm{x}}_i$ via:
\begin{equation}\label{imatcher_equation_warping}
    \hat{\bm{x}}_i = \hat{\mathbf{R}} \bm{x}_i + \hat{\mathbf{t}}, \quad \forall i = 1, \ldots, N
\end{equation}
where $\hat{\mathbf{R}} \in \mathit{SO}(3)$ and $\hat{\mathbf{t}} \in \mathbb{R}^3$ are the estimated rotation matrix and translation vector, respectively. The transformed point cloud is then defined as $\hat{\bm{X}} = \{ \hat{\bm{x}}_i \}_{i=1}^N$. This pre-alignment allows us to construct the set of correspondences by searching for the nearest neighbors in 3D space.
$\bm{y}_{\tilde{i}} \in \bm{Y}$, the target correspondent point of $\bm{x}_i \in \bm{X}$, is defined as follows: 
\begin{equation}\label{imatcher_eq_matching_repos}
\bm{y}_{\tilde{i}} = \underset{\bm{y}_j \in \bm{Y}}{\arg\min} \big( \| \hat{\bm{x}}_i - \bm{y}_j \| \big) ~~;\hat{\bm{x}}_i \in  \hat{\bm{X}}
\end{equation}

The features of the query source point and its target correspondent are then concatenated. The resulting feature vector:
\begin{equation}\label{imatcher_gcnn_corres_cat}
\hat{\bm{h}}^{X \leftarrow Y}_i \coloneqq \left[ \hat{\bm{h}}^X_i, \hat{\bm{h}}^Y_{\tilde{i}} \right] \in \mathbb{R}^{2d}
\end{equation}
$\hat{\bm{h}}^{X \leftarrow Y}_i \in \mathbb{R}^{2d}$ is then projected back to the original dimension $d$ using a linear layer. Let $\mathbf{W} \in \mathbb{R}^{d \times 2d}$ and $\mathbf{b} \in \mathbb{R}^d$ be the weight matrix and bias vector of a learnable linear transformation. The projected feature vector is given by:
\begin{equation}\label{imatcher_linear_proj}
\bm{h}^{X \leftarrow Y}_i \coloneqq \mathbf{W} \hat{\bm{h}}^{X \leftarrow Y}_i + \mathbf{b} \in \mathbb{R}^d
\end{equation}
The features ${\bm{h}}^{Y \leftarrow X}_j$ are obtained in the same way for points in $\bm{Y}$.

~\cref{imatcher_tab_ablation} from the ablation study shows that this step provides a significant performance improvement over a simple top-score selection from $\hat{\bm{S}}$. During the training phase, we utilize the ground-truth rigid transformation to warp the point clouds, preventing potential divergences caused by feature ambiguity in the early epochs of the learning process. \\
%
% --------

\textbf{1$^{st}$ Order spatial compatibility-based confidence score learning}\label{imatcher_subsub_matchabili}

Once the source-to-target and target-to-source matching is performed thanks to the previous scheme discussed in~\cref{imatcher_subsub_repos}, we propose to predict a matchability score for each source and target point based on the widely used First-Order Spatial Compatibility (FOSC) measures~\cite{lee2021deep,chen2022sc2}.

Let us note $\bm{\mathcal{C}}^{xy}$ with $\bm{\mathcal{C}}^{xy} = \{ (\bm{x}_i,\bm{y}_{\tilde{i}})~~;~~{\forall~ 1\leq i\leq  M   \}}$  the set of source-to-target matching pairs. The relative deviations between each query pair $(\bm{x}_i,\bm{y}_{\tilde{i}})$ and all the other $ (\bm{x}_j,\bm{y}_{\tilde{j}})$ are estimated, with $j$ browsing all the indices in $\bm{\mathcal{C}}^{xy}$, to build a geometric-aware spatial matrix $\bm{G}^X \in \mathbb{R}^{M \times M}$ with: 
\begin{equation}
     \bm{G}^X_{ij} =\lvert  d^X_{ij} - d^{Y}_{\tilde{i}\tilde{j}} \rvert
\end{equation}
with the distances $d_{ij}$ defined as: 
\begin{equation}
     d^X_{ij} = \lvert \bm{x}_i  -  \bm{x}_j \rvert ~~ \text{and} ~~
     d^{Y}_{\tilde{i}\tilde{j}} = \lvert \bm{y}_{\tilde{i}} -  \bm{y}_{\tilde{j}} \rvert
\end{equation}

In ideal rigid registration scenarios (\textit{i.e.}, with complete-to-complete matching and noise-free point clouds), both \((\bm{x}_i,\bm{y}_{\tilde{i}})\) and \((\bm{x}_j,\bm{y}_{\tilde{j}})\) are inlier correspondences if and only if \(\bm{G}^X_{ij} = 0\).  However, real-world application implies the presence of noise and occlusions. Thus, this exact equality rarely holds. Nevertheless, $\bm{G}^X_{ij}$ still tends to zero for inlier correspondences. To accommodate these imperfections, we introduce a soft compatibility matrix:
\begin{equation}
     \hat{\bm{G}}^X_{ij} = {\sigma}  \times \bm{G}^X_{ij} 
\end{equation}
where \(\sigma\) is a learned temperature to regulate the influence of perturbations and helps mitigate the impact of occlusions and noise, which affect measurement reliability. An exponential function is then applied to the softened compatibility matrix to get an intermediate confidence matrix $\tilde{\bm{G}}^X \in \mathbb{R}^{M \times M}$.
\begin{equation}
     \tilde{\bm{G}}^X_{ij} =  \exp{\big(-\hat{\bm{G}}^X_{ij}\big)}
\end{equation}

The distance consistency of the rigid transformation ensures that inlier correspondences have values close to $1$ in clean, occlusion-free point clouds, as all $\bm{G}^X_{ij}$ approaches zero. To obtain a single score $\alpha^X_i$ for each point $\bm{x}_i$, we compute the mean over all other correspondences:
\begin{equation}\label{imatcher_alpha_mean}
     \alpha^X_i =\sum_j^M  \frac{{\tilde{\bm{G}}}^X_{ij}}{M} 
\end{equation}

\begin{figure}[!h]
\centerline{\includegraphics[width=1\columnwidth]{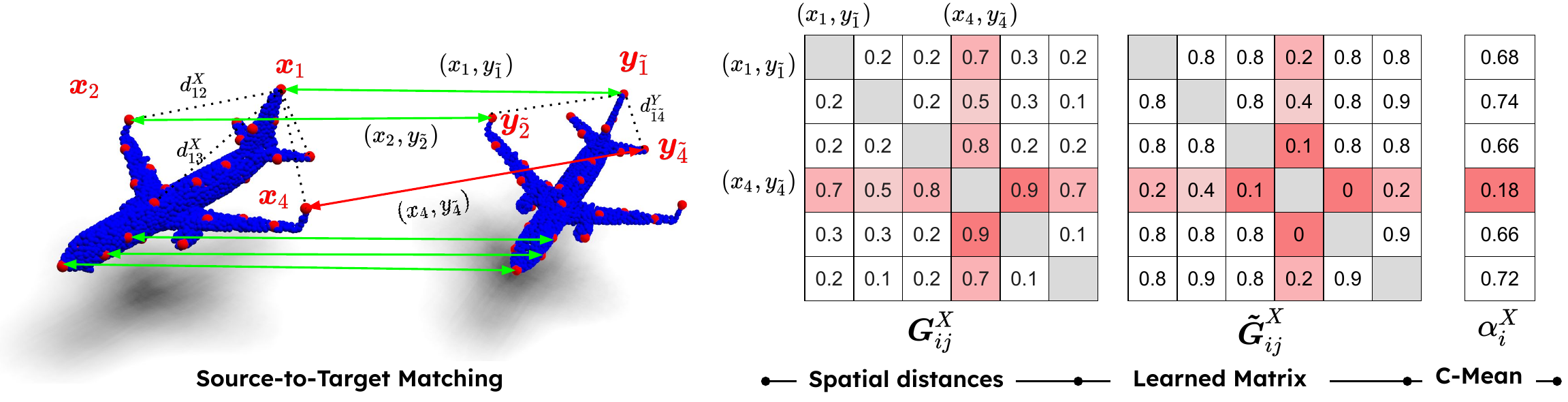}}
\caption{Geometric Global Consistency illustration with dummy values. Here all the source-to-target correspondences are inliers except for the $\bm{\mathcal{C}}^{xy}_4 =  (\bm{x}_4,\bm{y}_{\tilde4})$. \textbf{C-Mean} is for column-wise mean.}
\label{fig_fosc}
\end{figure}

The previous steps are mapped in~\cref{fig_fosc}. It shows the initial spatial distances $\bm{G}^X_{ij}$ for $6$ source-to-target correspondences and their projection into single scores $\alpha^X$, where all are inliers except the $4^{th}$ pair. The first matrix highlights how the $4^{th}$ row and column exhibit significantly higher deviations, indicating the outlier. This effect carries over to the final matchability scores $\alpha^X_i$ (last column), where most inliers score above $0.66$, while the outlier ($\alpha^X_4$) drops to $0.18$, confirming its unreliability.

To further limit noise and occlusion effects and maintain robustness in real-world settings, $\alpha^X_i$  
is projected into a higher-dimensional space $d$:
\begin{equation}
\hat{\bm{v}}^X_i = \bm{h}^2_\theta (\alpha^X_i) ~~;~~ \hat{\bm{v}}^X_i \in  \mathbb{R}^{d}
\end{equation}
where $\bm{h}^2_\theta$ is composed of three linear projections followed by \textit{Batch Normalization} and \textit{ReLU} functions, 
with dimensions equal to $\frac{d}{4}$, $\frac{d}{2}$, and $d$, respectively. These operations are employed in the same way to get the target point cloud  $\hat{\bm{v}}^Y_j$ for all the points $\bm{y}_j \in \bm{Y}$.
%
% -------------
\subsection{Local and global information fusion for final assignment matrix estimation} \label{imatcher_final_matrix}
The last module of \textit{iMatcher} aims to refine the score matrix ${\hat{\bm{S}}} \in \mathbb{R}^{M\times N}$ for use in rigid transformation estimation. To do so, we exploit both the geometric consistency encoding vector $\hat{\bm{v}}^X_i$ and the local graph feature vector ${\bm{h}}^{X \leftarrow Y}_i$. First, the two vectors are concatenated to a single vector  ${\bm{v}^X_i} \in \mathbb{R}^{2d}$, which is then fed to the matchability learning module that predicts the final confidence score as follows:
\begin{equation}
\bm{v}^X_i \coloneqq \left[ \hat{\bm{v}}^X_i, \bm{h}^{X \leftarrow Y}_i \right] \in \mathbb{R}^{2d}
\end{equation}
\begin{equation}\label{imatcher_eq_tau}
\tau^X_i = \text{Sigmoid}\Big(\bm{h}^3_\theta(\bm{v}^X_i)\Big)
\end{equation}
where $\bm{h}^3_\theta$ consists of four linear layers with dimensions $\frac{d}{4},\frac{d}{4},\frac{d}{2}$ and $1$.

We then define a \textit{matchability matrix} $\bm{S}^{\mathcal{M}}_{ij} \in [0,1]^{M \times N}$ as the outer product:
\begin{equation}\label{imatcher_eq_tau_matrice}
\bm{S}^{M}_{ij} = \tau^X \cdot (\tau^Y)^\top \quad \text{so that} \quad \bm{S}^{M}_{ij} =  \tau^X_i \cdot \tau^Y_j
\end{equation}

The final score matrix ${\bm{S}}$ is then defined as: 
\begin{equation}\label{imatcher_eq_softmax}
\bm{S} = \bm{S}^{\mathcal{M}} \circ \left( \text{Softmax}_{\text{row}}(\hat{\bm{S}}) \circ \text{Softmax}_{\text{col}}(\hat{\bm{S}}) \right)
\end{equation}
where $\circ$ is the element-wise product, and the \textit{dual-softmax} operations are defined as follows:
\[
\text{Softmax}_{\text{row}}(\hat{\bm{S}}_{ij}) = \frac{\exp(\hat{S}_{ij})}{\sum_{k=1}^N \exp(\hat{S}_{ik})}, \quad
\text{Softmax}_{\text{col}}(\hat{\bm{S}}_{ij}) = \frac{\exp(\hat{S}_{ij})}{\sum_{k=1}^M \exp(\hat{S}_{kj})}
\]
%
% --------------
\section{Experimental Validation} \label{sec.expe}
To evaluate the performance and versatility of the proposed \textit{iMatcher}, we perform extensive studies on three different types of data: (i) outdoor point clouds (KITTI~\cite{kitti} and KITTI-360~\cite{kitti360}), (ii) indoor scans (3DMatch~\cite{zeng20173dmatch}), and (iii) object-centric point clouds (TUD-L~\cite{hodan2018bop} and MVP-RG~\cite{pan2024robust}). The following sections provide a detailed breakdown of each experiment. All implementation details and parameter configurations are described in~\cref{imatcher_implementation} above. 

% \subsection{Datasets} \label{sec.datasets}

% \subsection{Metrics} \label{sec.metrics}

%
% -------------
\subsection{Implementation Details}\label{imatcher_implementation}
The method is implemented in PyTorch and trained on an NVIDIA RTX A5000 GPU. The following table~\ref{tab.implement} details the learning parameters (i.e. Number of epochs (\textbf{Ep}), batch size (\textbf{B.S}), optimizer (\textbf{Opt.}), the feature dimension $d$, the number of neighbors in the graph construction $k$ and the minimized loss. All the experiments use a $10^{-4}$ learning rate with an exponential decay (every single epoch on 3DMatch and every $4$ epochs on KITTI) as in~\cite{geotransformer}.
\begin{table}[!h]
\scriptsize
    \centering
    \caption{\textit{iMatcher} Training parameters for the used datasets. } \label{tab.implement}
    \begin{tabularx}{\linewidth}{l|l|l|l|l|l|X}
        \hline

        \hline

        \hline

        \textbf{Dataset} & \textbf{Ep} & \textbf{B.S} & \textbf{Opt.}  & \textbf{$d$} & \textbf{$k$} & \textbf{Loss} \\
        \hline
        3DMatch & 40 & 1 & Adam &  256 & 12 & \tiny{($L_{oc}+L_m$)}~\cite{geotransformer}  \\
        KITTI & 40 & 1 & Adam &  256 & 32 & \tiny{($L_{oc}+L_m$)}~\cite{geotransformer}  \\
        TUD-L (DCP) & 20 & 16 & Adam &  256 & 12 & \tiny Gap Loss~\cite{mdgat} \\
        TUD-L (RPMNet) & 30 & 16 & Adam &  96 & 12 & \tiny Gap Loss~\cite{mdgat}  \\
        MVP-RG & 200 & 1 & Adam &  256 & 12 & \tiny Gap Loss~\cite{mdgat}  \\
        \hline

        \hline

        \hline
        \end{tabularx}
\end{table}
%
% ------------------
\subsection{Evaluation on Outdoor Benchmark}\label{imatcher_subsect_kitti}
% -----------------
%
For this outdoor odometry challenge evaluation, we conduct two experiments. The first replicates the experimental setup from~\cite{geotransformer, huang2021predator, bai2020d3feat}. It reports the performance of various state-of-the-art (SOTA) methods using Registration Recall (RR, in \%), Relative Rotation Error (RRE, in degrees), and Relative Translation Error (RTE, in cm). Next, following~\cite{shi2023rdmnet}, we assess the generalization ability of \textit{iMatcher} by validating a model trained on the KITTI dataset with the KITTI-360 dataset using the same metrics. For both datasets, we adopt GeoTransformer~\cite{geotransformer} as our baseline. This method employs a coarse-to-fine registration approach in two stages: superpoint and dense point matching. Our study focuses on dense point matching, as it relies on a differentiable Sinkhorn algorithm~\cite{geotransformer} to compute the similarity score matrix. We propose replacing this module with \textit{iMatcher}. Let us note $\boldsymbol{p}_i \in \mathbb{R}^3$  and  $\boldsymbol{q}_i\in \mathbb{R}^3$ two paired superpoints,  $\boldsymbol{X}^p_i \in \mathbb{R}^{M\times3}$ and $\boldsymbol{Y}^q_i \in \mathbb{R}^{N\times3}$ the sample of points around them and their associated features $\boldsymbol{F}^p_i \in \mathbb{R}^{M\times D}$ and $\boldsymbol{F}^q_i  \in \mathbb{R}^{N\times D}$, respectively. The score matrix $\boldsymbol{S}^{p_iq_i} \in \mathbb{R}^{M\times N} $ associated with the pair $\boldsymbol{p_i}$ and $\boldsymbol{q_i}$ is given by: 
\begin{equation}
    \boldsymbol{S}^{p_iq_i} = \text{\textit{iM}} \Big(\boldsymbol{X}^p_i,\boldsymbol{Y}^q_i,\boldsymbol{F}^p_i ,\boldsymbol{F}^q_i \Big)
\end{equation}
where \text{\textit{iM}} is the \textit{iMatcher} function described in~\cref{sec:method}, where $\boldsymbol{X}$, $\boldsymbol{Y}$, $\boldsymbol{F}^X$ and $\boldsymbol{F}^Y$ are replaced by $\boldsymbol{X}^p_i$, $\boldsymbol{Y}^q_i$, $\boldsymbol{F}^p_i$ and $\boldsymbol{F}^q_i$, respectively. This operation is repeated for all paired superpoints, and the final dense correspondences set $C$ is given by the union of the largest $k$ scores of each $\boldsymbol{S}^{p_iq_i}$: 
\begin{equation} \label{eq.matching.geotr}
    \footnotesize
    \bm{\mathcal{C}}^{xy} = \bigcup_{i=1}^{N_m} \Bigg(\Big\{(\boldsymbol{x}_m,\boldsymbol{y}_n) \in  \boldsymbol{F}^p_i \times \boldsymbol{F}^q_i| {S}^{p_iq_i}_{m,n} \in \text{mutual\_top}_k(\boldsymbol{S}^{p_iq_i}) \Big\}\Bigg)
\end{equation}
where $N_m$ is the number of superpoint correspondences. The remaining modules of the GeoTransformer pipeline are kept unchanged.
\subsubsection{KITTI}

This odometry dataset consists of LiDAR scans collected from a moving vehicle. Following~\cite{geotransformer, huang2021predator}, we refine the ground truth poses using ICP and evaluate only on point cloud pairs separated by at least 10 meters. As in~\cite{geotransformer, huang2021predator}, we train \textit{iMatcher} on sequences $0$-$5$, validate on sequences $6$-$7$, and use sequences $8$-$10$ for testing.~\cref{tab.KITTI.RR_rte_rre} demonstrates that \textit{iMatcher} significantly improves upon the GeoTransformer baseline by reducing the RTE by approximately 19\% and the RRE by 13\%, securing the runner-up position, closely trailing RDMNet on these metrics.~\cref{tab.kitti.numcorr} compares matching performance, highlighting \textit{iMatcher}'s substantial impact on dense point correspondences, with an inlier ratio improvement ranging from 10.2\% to 19.4\% for 250 to 5000 putative matches.~\cref{fig.kitti.matching} provides a qualitative comparison of matches obtained with \textit{iMatcher}, GeoTransformer~\cite{geotransformer}, and RDMNet~\cite{shi2023rdmnet} on selected evaluation samples.

\begin{table}[!h]\scriptsize
\centering
% \scriptsize % Makes the table more compact
\caption{Registration performances on KITTI. GeoT-V2 is a modified version of GeoTransformer~\cite{geotransformer} on which we replaced the transformer by the 3DRoformer from RDMNet~\cite{shi2023rdmnet} }
%\renewcommand{\arraystretch}{1.1}  % Adjust row height for better readability

% \begin{tabularx}{\linewidth}{l l| X c c}
\begin{tabularx}{\linewidth}{l l| X X X}

\hline

\hline

\hline
\textbf{Method} & \textbf{Ref} & RTE (cm) $\downarrow$ & RRE (deg) $\downarrow$ & RR~(\%)$\uparrow$ \\ 
\hline

% Predator~\cite{huang2021predator}  & CVPR'19 & 6.8  & 0.27 & 90.6  \\
% SpinNet~\cite{SpinNet}             & CVPR'21 & 9.9  & 0.47 & 99.1  \\
%CoFiNet~\cite{yu2021cofinet}       & NeurIPS'21 & 8.2 & 0.41 & \textbf{99.8} \\
RoCNet++~\cite{slimani2024rocnet++} & PatRec'24 & 7.3 & 0.23 & \textbf{99.8} \\
% MAC~\cite{MAC}                     & CVPR'23  & 6.5  & 0.40 & 99.5  \\
RegFormer~\cite{liu2023regformer}  & ICCV'23  & 8.4  & 0.24 & \textbf{99.8} \\
% PTT~\cite{pointree}                & TCSVT'25 & 6.3  & 0.23 & \textbf{99.8} \\
GPNF~\cite{zhu2024gpnf}            & ACCV'24  & 6.0  & 0.23 & \textbf{99.8} \\
PointDT~\cite{pointdt}             & HS'24    & 6.3  & {0.22} & \textbf{99.8} \\
DCATr~\cite{DCATr}                 & CVPR'24  & 6.6  & {0.22} & \underline{99.7} \\
SGNet~\cite{wu2024sgnet}           & IROS'24  & 5.4  & 0.24 & \textbf{99.8} \\
S2Reg~\cite{xu2025s2reg}           & PatRec'24 & 6.3 & 0.23 & \textbf{99.8} \\
PARE-Net~\cite{parenet}            & ECCV'24  & {5.1} & 0.23 & \textbf{99.8} \\
RDMNet~\cite{shi2023rdmnet}            & ITSS'23  & \underline{4.8}& \textbf{0.19} & \textbf{99.8} \\

\hline

{GeoT. (Sink.)}  & CVPR'22 & 6.2  & 0.23  & \textbf{99.8} \\
\textbf{GeoT. (\textit{iMatcher})}   &   & \underline{4.8} & \underline{0.20}  & \textbf{99.8} \\

 \rowcolor{blue!10} \textcolor{blue}{\emph{{Improvement ($\%$})} } &   & \textit{\textcolor{blue}{+22.6}} & \textit{\textcolor{blue}{+13.0}} & \textit{\textcolor{blue}{+0.0}} \\

\hline

{GeoT.-V2 (Sink.)}  &   & 6.5  & 0.23  & \textbf{99.8} \\
% [32m[2025-02-09 10:42:51][0m [1;30m[CRIT][0m [1;31m  Registration, RR: 0.998, RRE: 0.229, RTE: 0.065[0m

\textbf{GeoT.-V2 (\textit{iMatcher})}   &   & \textbf{4.7}  & \underline{0.20}  & \textbf{99.8} \\

\rowcolor{blue!10} \textcolor{blue}{\emph{{Improvement ($\%$})} }  &   & \textit{\textcolor{blue}{+27.7}} & \textit{\textcolor{blue}{+13.0}} & \textit{\textcolor{blue}{+0.0}} \\

% Registration, RR: 0.998, RRE: 0.195, RTE: 0.047
\hline

\hline

\hline

\end{tabularx}
\label{tab.KITTI.RR_rte_rre}
\end{table}

\begin{figure*}
\centerline{\includegraphics[width=1\columnwidth]{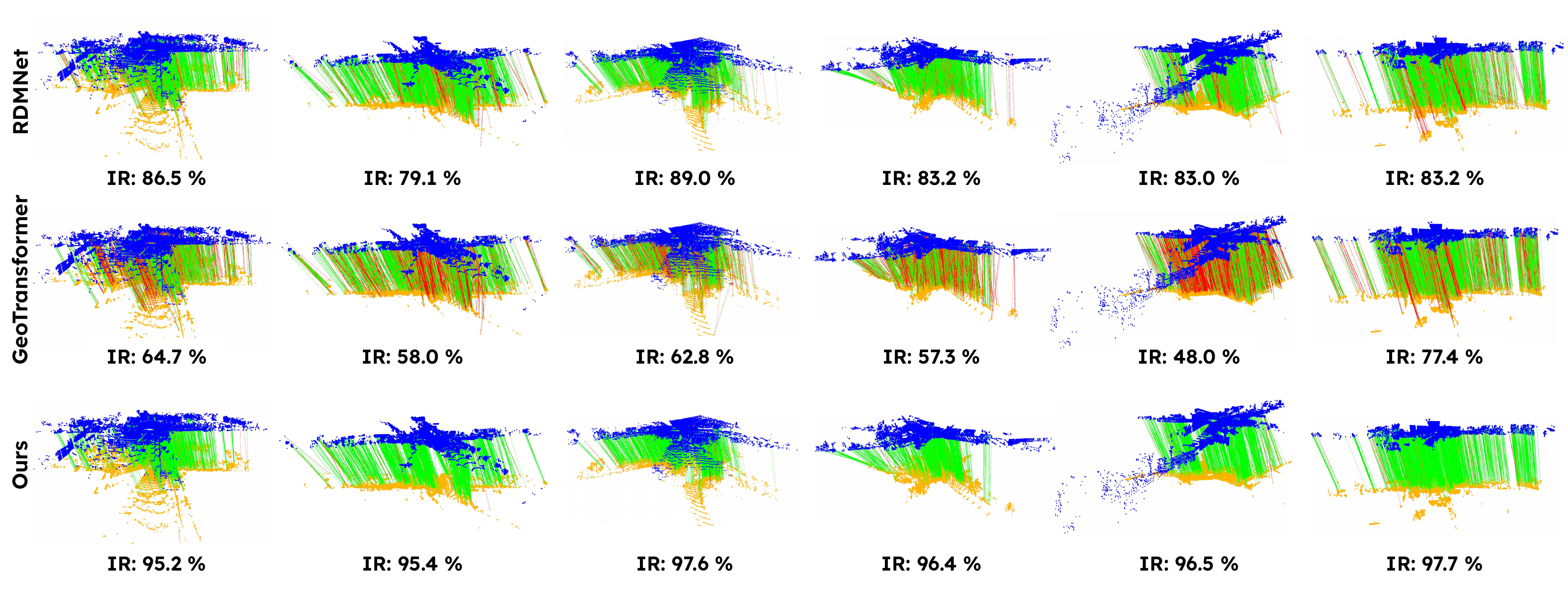}}
\caption{Predicted matches by RDMNet (top), GeoTransformer-Sinkhorn (middle), and GeoTransformer-\textit{iMatcher} (bottom) on KITTI.}
\label{fig.kitti.matching}
\end{figure*}

%
%----
%

\small
\begin{table}[!h]\scriptsize
% \footnotesize  % Makes the table more compact
\caption{Performances on KITTI under various numbers of correspondences.}

\begin{center}
\begin{tabularx}{\linewidth}{l l | X  X X}
\hline

\hline

\multicolumn{5}{c}{Feature Matching Recall{ ($\uparrow$)}} \\
\hline
\textbf{Method} & \textbf{Ref}&  \textbf{5000} & \textbf{1000} & \textbf{250} \\
\hline
RDMNet~\cite{shi2023rdmnet} &  ITSS'23 & \textbf{99.8} & \textbf{99.8} & \textbf{99.8} \\
PARE-Net~\cite{parenet} &  ECCV'24& \textbf{99.8} & \textbf{99.8} & \textbf{99.8} \\

\hline

{GeoT. } & CVPR'22 & \textbf{99.8} & \textbf{99.8} & \textbf{99.8} \\
\textbf{GeoT.  (\textit{iMatcher})} &  - & \textbf{99.8} & \textbf{99.8} & \textbf{99.8} \\
\hline

\hline

\multicolumn{5}{c}{Inlier Ratio{ ($\uparrow$)}} \\
\hline
RDMNet~\cite{shi2023rdmnet}& ITSS'23 &  86.7 & 93.0  &95.3 \\

PARE-Net~\cite{parenet} &  ECCV'24 & 59.7 & 70.5  &  72.4 \\

\hline
% RDMNet & 86.7 & 93.0 & 95.3 \\
{GeoT. (Sink.)} &   CVPR'22 & 75.7 & 86.0 & 87.5 \\
\textbf{GeoT.  (\textit{iMatcher})}& -  & \textbf{95.1} & \textbf{97.2} & \textbf{97.7} \\
\rowcolor{blue!10} \textcolor{blue}{\emph{{Improvement} ($\%$)} } &   & \textit{\textcolor{blue}{+25.6}} & \textit{\textcolor{blue}{+13.0}} & \textit{\textcolor{blue}{+11.7}} \\

\hline 

\hline 

\hline
\end{tabularx}
\label{tab.kitti.numcorr}
\end{center}
\end{table}
\normalsize

% ------
\subsubsection{KITTI-360}
Inspired by~\cite{shi2023rdmnet}, we evaluate our method on KITTI-360 using the pre-trained weights from KITTI. KITTI-360 is another outdoor dataset that provides more precise and more accurate ground-truth poses. For this evaluation, we use sequences $0$ to $10$, excluding $1$ and $8$, resulting in 5,888 point cloud pairs.~\cref{tab.KITTI-360.num_corr} shows that \textit{iMatcher} achieves a relative improvement in the inlier ratio over the baseline by 13.3\% to 28\%, making it the best-performing model under this metric, regardless of the number of correspondences used. However, it results in a slight decrease in Feature Matching Recall (FMR), with a negligible drop of 0.0003\%. Regarding pose estimation,~\cref{tab.KITTI-360.ransac} highlights a significant performance boost, with improvements of 26\% in RRE and 31\% in RTE. Notably, RDMNet achieves the best overall results across the three evaluated metrics.

\small
\begin{table}[!h]\scriptsize
% \footnotesize  % Makes the table more compact
\caption{Performances on KITTI-360 Dataset under various numbers of correspondences.}
\begin{center}
\begin{tabularx}{\linewidth}{l l | X X X | X X X}
\hline

\hline

\hline

\multicolumn{2}{c|}{} & \multicolumn{3}{c|}{{FMR($\uparrow$)}} & \multicolumn{3}{c}{{IR ($\uparrow$)}} \\
\hline
\textbf{Method} & \textbf{Ref} & \textbf{5000} & \textbf{1000} & \textbf{250} & \textbf{5000} & \textbf{1000} & \textbf{250} \\
\hline
RDMNet~\cite{shi2023rdmnet} & ITSS'23 & \textbf{99.95} & \textbf{99.95} & \textbf{99.97} & 84.0 & 91.0 & 93.7 \\ 
PARE-Net~\cite{parenet} & ECCV'24 & {99.57} & {99.58} & {99.66} & 66.7 & 66.7 & 69.3 \\
\hline
GeoT.  (Sink.) & CVPR'22 & \underline{99.92} & \underline{99.93} & \underline{99.95} & 73.6 & 84.0 & 85.9 \\
\textbf{GeoT.  (\textit{iMatcher})} & & \underline{99.92} & 99.92 & 99.92 & \textbf{94.2} & \textbf{96.6} & \textbf{97.3} \\
\rowcolor{blue!10} \textcolor{blue}{\emph{{Improvement} ($\%$)} } & & \textit{\textcolor{blue}{+0.00}} & \textit{\textcolor{black}{-0.0001}} & \textit{\textcolor{black}{-0.0003}} & \textit{\textcolor{blue}{+28.0}} & \textit{\textcolor{blue}{+15.0}} & \textit{\textcolor{blue}{+13.3}} \\
\hline

\hline

\hline
\end{tabularx}
\label{tab.KITTI-360.num_corr}
\end{center}
\end{table}
\normalsize

\begin{table}[!h]\scriptsize
\caption{Registration performances on KITTI-360.In addition to estimating the pose using local-to-global registration with GeoTransformer (both with Sinkhorn and iMatcher), we also report results where the pose is estimated using RANSAC.}
\centering
\begin{tabularx}{\linewidth}{l l | X X X}
\hline

\hline

\hline

\textbf{Method} & \textbf{Ref} & RTE (cm)~$\downarrow$ & RRE (deg)~$\downarrow$ & RR (\%)~$\uparrow$ \\
\hline
RDMNet~\cite{shi2023rdmnet} & ITSS'23 & \textbf{7.9} & \textbf{0.23} & \textbf{99.93} \\
PARE-Net~\cite{parenet} & ECCV'24 & 9.1 & 0.28 & 99.11 \\
\hline
\multicolumn{5}{c}{\textbf{iMatcher in GeoTransformer}} \\
\hline
\multicolumn{1}{l}{\textbf{Desriptor}} & \textbf{Matching / Pose Est.} & RTE (cm)~$\downarrow$ & RRE (deg)~$\downarrow$ & RR (\%)~$\uparrow$ \\
\hline
GeoTransformer & Sinkhorn / RANSAC & 13.5 & 0.431 & 99.49 \\
GeoTransformer  & \textbf{iMatcher} / RANSAC & 10.0 & 0.299 & 99.81 \\
\rowcolor{blue!10} \textcolor{blue}{\emph{Improvement (\%)}} & & \textit{\textcolor{blue}{+25.9}} & \textit{\textcolor{blue}{+30.6}} & \textit{\textcolor{blue}{+0.3}} \\

\hline
GeoTransformer & Sinkhorn / LGR & 9.1 & \underline{0.247} & \underline{99.9} \\
GeoTransformer  & \textbf{iMatcher} / LGR & \underline{8.5} & 0.303 & \underline{99.9} \\
\rowcolor{blue!10} \textcolor{blue}{\emph{Improvement (\%)}} & & \textit{\textcolor{blue}{+6.6}} & \textit{\textcolor{black}{-22.7}} & \textit{\textcolor{blue}{+0.0}} \\
\hline

\hline

\hline

\end{tabularx}

\label{tab.KITTI-360.ransac}
\end{table}

%------
\subsection{Evaluation on Indoor Benchmark}
As in the previous experiment, we choose GeoTransformer~\cite{geotransformer} as our baseline and use its features for the dense point assignment using \textit{iMatcher}. The same methodology as in~\cref{imatcher_subsect_kitti} is used here.
\subsubsection{3DMatch} 
This dataset consists of RGB-D scans of indoor environments, divided into 46 scenes for training, 8 for validation, and 8 for testing. Following the experiment setup in~\cite{geotransformer}, we use the preprocessed point clouds from Predator~\cite{huang2021predator} and evaluate our model under the same metrics as~\cite{geotransformer}.~\cref{tab.3dmatch} shows that \textit{iMatcher} improves all metrics related to pose estimation (RR, RRE, and RTE) and correspondence prediction (IR and FMR). Notably, our method achieves the highest IR score, surpassing the second-best method, DCATr~\cite{DCATr}, by a significant margin of 6.2\%. It ranks second in FMR, RRE, and RTE, trailing DCATr, while securing the third position in RR, where Diff-Reg~\cite{wu2024diff} attains the best result.

\begin{figure*}
\centerline{\includegraphics[width=1.\columnwidth]{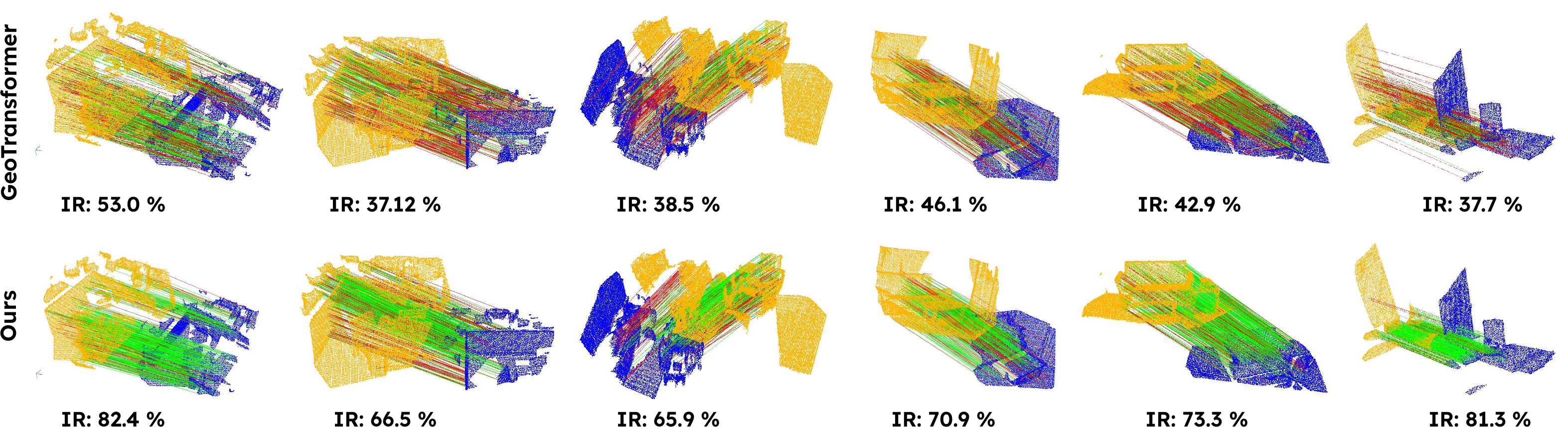}}
\caption{Putative correspondences predicted by GeoTransformer-Sinkhorn (top) and GeoTransformer-\textit{iMatcher} (bottom) on 3DMatch. }
\label{fig.kitti.matching}
\end{figure*}

\begin{table}[!h]\scriptsize
    % \scriptsize  % Makes the table more compact
    \caption{Performance on the 3DMatch dataset.}

    \centering
    \begin{tabularx}{\columnwidth}{l l | X X X X X}
    
    \hline 
    
    \hline 
    
    \hline

        % \toprule
        \textbf{Method} & \textbf{Ref} & RR{ ($\uparrow$)}  & IR{ ($\uparrow$)}  & FMR{ ($\uparrow$)}  & RRE{ ($\downarrow$)} & RTE{ ($\downarrow$)} \\
        \midrule
        % FCGF \cite{choy2019fully} & ICCV'19 & 83.3 & 48.7 & 97.0 & 1.949 & 6.6 \\
        D3Feat \cite{bai2020d3feat} & CVPR'20 & 83.4 & 40.4 & 94.5 & 2.161 & 6.7 \\
        Predator \cite{huang2021predator}& CVPR'21 & 90.6 & 57.1 & 96.5 & 2.029 & 6.4 \\
        CoFiNet \cite{yu2021cofinet} & NeurIPS'21 & 91.5 & 61.9 & 98.1 & 2.011 & 6.1 \\        
        {DCATr~\cite{DCATr}}& CVPR'24 & \underline{92.1} & 75.0 & \textbf{98.1} & \textbf{1.536} & \textbf{5.0} \\
        Diff-Reg~\cite{wu2024diff} & ECCV'24 & \textbf{95.0} & 30.9 & 96.3 &  -  &  -  \\

        \hline
        GeoT. (Sink.) \cite{geotransformer}&  CVPR'22 & 91.5 & 70.3 & 97.7 & 1.625 & 5.3 \\
        \textbf{GeoT. (\textit{iMatcher})} & & {91.7} & \textbf{81.2} & \underline{97.9} & \underline{1.591} & \underline{5.1} \\
        \rowcolor{blue!10} \textcolor{blue}{\emph{{Improvement ($\%$})} }  &   & \textit{\textcolor{blue}{+0.0}}  & \textit{\textcolor{blue}{+15.5}}  & \textit{\textcolor{blue}{+0.0}} &  \textit{\textcolor{blue}{+2.1}} & \textit{\textcolor{blue}{+3.8}} \\

        % \bottomrule
    \hline 

    \hline 
    
    \hline

    \end{tabularx}\label{tab.3dmatch}

\end{table}

% ------------------------
\subsection{Object-centric Registration}
In addition to indoor and outdoor experiments, \textit{iMatcher} is also evaluated on object-centric registration. For this purpose, we employ two distinct datasets: TUD-L~\cite{hodan2018bop} and MCP-RG~\cite{pan2024robust}. In this setting, the point cloud density is lower than that of KITTI and 3DMatch, which influenced our choice of state-of-the-art architectures, LoGDesc~\cite{slimani2024logdesc}, DCP~\cite{dcp}, and RPMNet~\cite{yew2020rpm} as baselines, eliminating the need for a fine-to-coarse matching process. The algorithm described in~\cref{sec:method} is applied straightforwardly, with feature matrices $\boldsymbol{F}^X \in \mathbb{R}^{M\times d}$ and $\boldsymbol{F}^Y \in \mathbb{R}^{N\times d}$ extracted using the selected models. To estimate the rigid transformation, we first binarize the predicted soft score matrix from iMatcher using a mutual (source-to-target and target-to-source) top-1 score selection.

\subsubsection{TUD-L: 6D object pose estimation}~\label{imacther_tudl} 
\begin{figure}[!h]
\centerline{\includegraphics[width=1.\columnwidth]{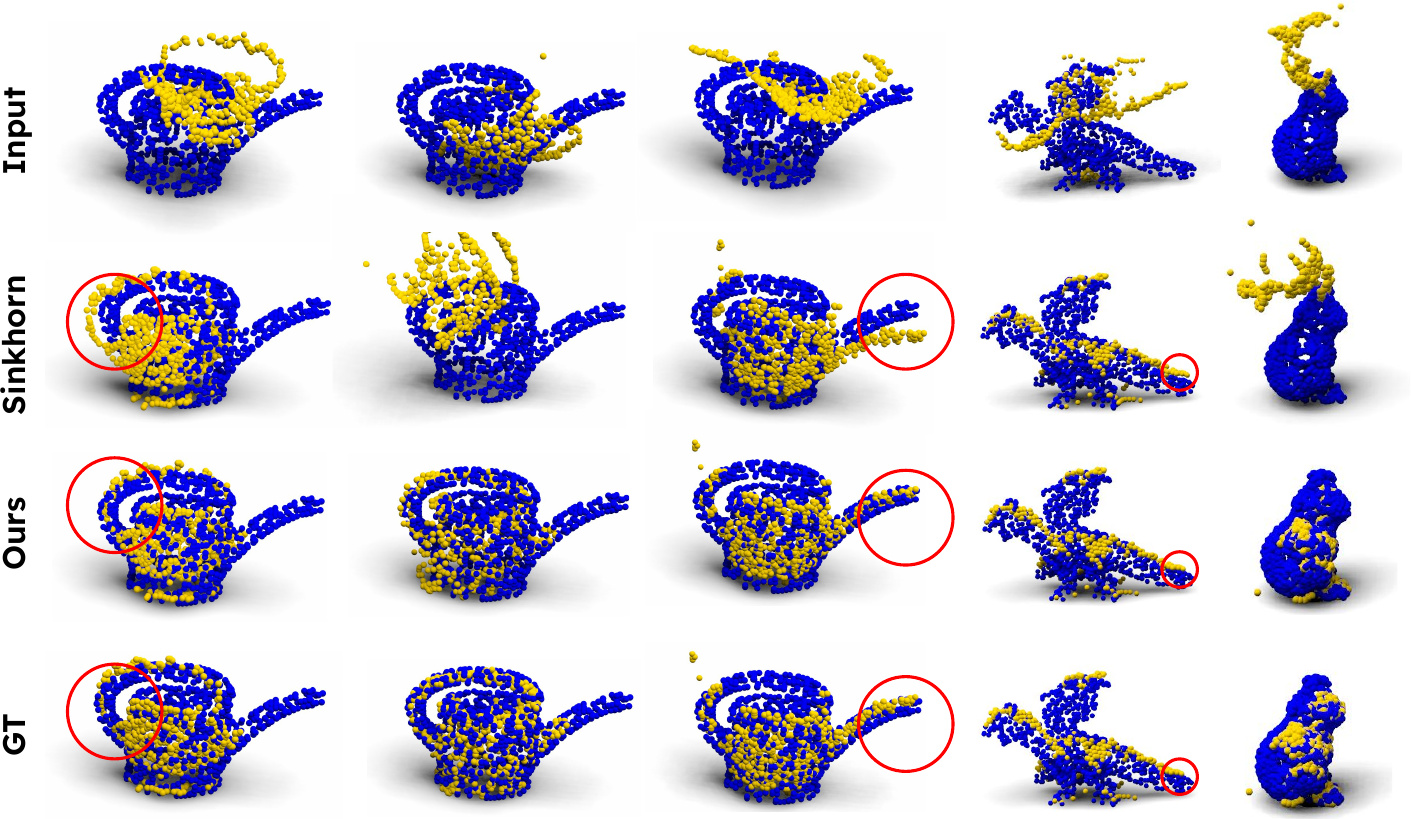}}
\caption{Illustration of performed registration by the two versions of DCP (\textit{i.e.}, with Sinkhorn and with \textit{iMatcher}) on the three classes of TUD-L.}
\label{fig.tudl}
\end{figure}

The TU Dresden Light (TUD-L) dataset serves as a challenging benchmark, consisting of captured scans of three household objects in motion. This evaluation focuses on registering the acquired point clouds with complete object models, a difficult task due to the high occlusion ratio between source and target data. Results from~\cite{jiang2024se}, summarized in Table~\ref{tab.tudl}, reveal that most learning-based models struggle with such conditions, leading to frequent registration failures. However, DCP~\cite{dcp} and RPMNet~\cite{yew2020rpm} demonstrate strong performance when integrated with the diffusion scheme from~\cite{jiang2024se}. Results from~\cite{jiang2024se}, summarized in~\cref{tab.tudl}, reveal that most learning-based models struggle with such conditions, resulting in frequent registration failures. However, DCP~\cite{dcp} and RPMNet~\cite{yew2020rpm} demonstrate strong performance when integrated with the diffusion scheme from~\cite{jiang2024se}. Inspired by this observation, we integrate \textit{iMatcher} into both models and compare its performance against a variant where Sinkhorn replaces it. Following~\cite{jiang2024se}, we report the mean Average Precision (mAP) at two different rotation and translation thresholds.~\cref{tab.tudl} shows that \textit{iMatcher} consistently outperforms Sinkhorn in both rotation and translation across all thresholds when applied to DCP, achieving the highest rotation mAP(5°) and matching Diff-RPMNet in translation mAP(2cm). Notably, compared to the original DCP, \textit{iMatcher} yields significant improvements of 68\%, 35\%, 93\%, and 73\% for mAP(5$^\circ$), mAP(10$^\circ$), mAP(1cm), and mAP(2cm), respectively.~\cref{fig.tudl} presents qualitative examples of registration results for the three objects.

\begin{table}[!h]\scriptsize
\centering
% \scriptsize  % Makes the table more compact
\caption{Performance on the TUD-L dataset. mAP are computed on RRE and RTE.}
%\renewcommand{\arraystretch}{0.9}  % Adjust row height for better readability

% \begin{tabularx}{\linewidth}{ll|XXXX}
\begin{tabularx}{\linewidth}{ll|XXXX}
\hline

\hline

\hline

\multirow{2}{*}{\textbf{Method}} & \multirow{2}{*}{\textbf{Ref}} &  \multicolumn{4}{c}{mean Average Precision (mAP) $\uparrow$} \\ 
\cline{3-6}

% \cmidrule{3-6}
                    &    & 5° & 10° & 1cm & 2cm \\ 
\hline
%ICP~\cite{besl1992method}    &   SFIV'92  & 0.02 & 0.02 & 0.01 & 0.14 \\
%FGR~\cite{zhou2016fast}    &  Info. Sci.'16   & 0.00 & 0.01 & 0.04 & 0.25 \\
% TEASER~\cite{ref3}      & 0.13 & 0.17 & 0.03 & 0.22 \\
% FMR~\cite{huang2020feature}&  CVPR'20 & 0.02 & 0.09 & 0.02  & 0.06  \\
DCP~\cite{dcp}     &   ICCV'19 & 0.23 & 0.62 & 0.04 & 0.26 \\
RPMNet~\cite{yew2020rpm}   &  CVPR'20 & 0.42 & 0.33 & 0.20 & 0.50 \\
%IDAM~\cite{li2020iterative}   &   ECCV'20  & 0.03 & 0.05 & 0.02 & 0.08 \\
MN-IDAM ~\cite{dang2022learning}  &   ECCV'22  & 0.36 & 0.46 & 0.23 & 0.47 \\
MN-DCP~\cite{dang2022learning}&   ECCV'22 &0.70 &0.81 & 0.71 & 0.86 \\
{Diff-DCP}~\cite{jiang2024se}    &  NeuRIPS'24  & {0.65} & {0.85} & {0.73} & {0.94} \\
{Diff-RPMNet}~\cite{jiang2024se}    &   NeuRIPS'24   & \underline{0.90} & \textbf{0.98} & \textbf{0.98} & \textbf{0.99} \\   

\hline

RPMNet (Sink.)   &      & 0.87 & 0.96 & 0.95 & {0.97} \\ %ep30
% \textbf{RPMNet (Ours) }     & {0.87} &  {0.96} & {0.96} & {0.97} \\% epoch 24
\textbf{RPMNet (\textit{iMatcher}) }  &   & {0.89} &  {0.96} & {0.96} & {0.97} \\% epoch 329

\rowcolor{blue!10} \textcolor{blue}{\emph{{Improvement ($\%$)}}} &   & \textit{\textcolor{blue}{+2.3}}  & \textit{\textcolor{blue}{+0.0}}  & \textit{\textcolor{blue}{+1.1}} & \textit{\textcolor{blue}{+0.0}} \\
% mAP_R (5/10/20 degree): 0.866, 0.960, 0.982 | mAP_t (1/2/5 cm): 0.959, 0.970, 0.980 |   wo repos    , epoch24
% mAP_R (5/10/20 degree): 0.873, 0.963, 0.983 | mAP_t (1/2/5 cm): 0.959, 0.972, 0.982 | with repo   , epoch24

\hline

DCP (Sink.)     &    & 0.79 & 0.93 & 0.91 & {0.96} \\
\textbf{DCP (\textit{iMatcher}) }   &  & \textbf{0.91} & \underline{0.97} & \underline{0.97} & \textbf{0.99} \\ % 20 epochs
\rowcolor{blue!10} \textcolor{blue}{\emph{{Improvement ($\%$)}}} &   & \textit{\textcolor{blue}{+15.2}}  & \textit{\textcolor{blue}{+4.3}}  & \textit{\textcolor{blue}{+6.6}} & \textit{\textcolor{blue}{+3.1}} \\

% RPMNet~\cite{ref7}      & 0.42 & 0.33 & 0.20 & 0.50 \\
% \textbf{Diff-RPMNet}~\cite{jiang2024se}    & \textbf{0.90} & \textbf{0.98} & \textbf{0.99} & \textbf{0.98} \\   On pourrait ajouter ça

\hline

\hline

\hline

\end{tabularx} \label{tab.tudl}
\end{table}
%
% ---------
\subsubsection{MVP-RG: synthetic multi-view partial data}
The authors of~\cite{pan2024robust} introduced the Multi-View Partial virtual scan ReGistration (MVP-RG) dataset, a challenging benchmark for point cloud registration, derived from the multi-view virtual-scanned partial point cloud dataset MVP~\cite{pan2021variational}. To construct this dataset, they projected depth scans from multiple camera viewpoints onto image coordinates and subsampled them by retaining only the 2048 farthest points. Point cloud pairs were generated for each object based on a sufficient overlap criterion, resulting in 6400 pairs for training and 1200 pairs for evaluation. We compare iMatcher against Sinkhorn using LoGDesc features~\cite{slimani2024logdesc}. Following~\cite{slimani2024logdesc}, we select the 768 farthest points as input to mitigate the computational cost of the transformer module. As shown in~\cref{imatcher_tab_mvprg}, \textit{iMatcher} delivers significant improvements across all three evaluation metrics compared to Sinkhorn, achieving the best overall performance among state-of-the-art methods. It is only matched in translation error by an end-to-end PCR transformer~\cite{wang2024end}.
\begin{table}[!h]\scriptsize
% \scriptsize
\caption{{Performance on the MVP-RG dataset. }} 
\centering
\begin{tabularx}{\linewidth}{l l | X  |  X | X }
\hline

\hline

\hline

\textbf{Method} &  \textbf{Ref} & RRE $(\downarrow) $& RTE $(\downarrow)$ & \textbf{$L_{RMSE} (\downarrow)$}\\
\hline
%IDAM~\cite{li2020iterative} & ECCV'20  & 24.35$^\circ$ & 0.280 & 0.344 \\
% RGM~\cite{rgm} &  CVPR'21  & 41.27$^\circ$ & 0.425 & 0.583 \\
DCP~\cite{dcp}  & ICCV'19  & 30.37$^\circ$ & 0.273 & 0.634 \\
% DeepGMR & 43.74$^\circ$ & 0.353 & 0.608 \\
{Predator} \cite{huang2021predator} & CVPR'21 & {10.58}$^\circ$ & {0.067} & {0.125}  \\
RPMNet~\cite{yew2020rpm} & CVPR'20  & 22.20$^\circ$ &  {0.174} & 0.327 \\
GMCNet~\cite{pan2024robust} & RAL'24 & {16.57$^\circ$} & {0.174} & {0.246} \\
E2E PCR Tr. ~\cite{wang2024end} &  AIR'24 & 7.46$^\circ$ & \textbf{0.033}& \underline{0.097}  \\
\hline
% CCAG: end-to-end point cloud registration RAL’23   , 

{{LoGDesc  (Sink.)}}& ACCV'24 & {7.33$^\circ$}  &  \underline{0.043} &  {0.099}  \\  
% {{LoGDesc (FGR)}}& ACCV'24 & {9.08$^\circ$}  &  {0.061} &  {0.099}  \\  
% {{LoGDesc (FGR)*}}& -  & \underline{5.42$^\circ$}  &  {0.047} &  \textbf{0.073}  \\  

{\textbf{LoGDesc (\textit{iMatcher})}}& - &   \textbf{5.04}$^\circ$ &  \textbf{0.033} &  \textbf{0.073} \\  
\rowcolor{blue!10} \textcolor{blue}{\emph{{Improvement ($\%$)}}} &   & \textit{\textcolor{blue}{+31.2}}  & \textit{\textcolor{blue}{+23.3}}  & \textit{\textcolor{blue}{+26.3}} \\

\hline

\hline

\hline

\end{tabularx}

\label{imatcher_tab_mvprg}

\end{table}

\subsection{Time Consumption}

We further analyze the computational cost of \textit{iMatcher} in comparison to the Sinkhorn algorithm. Specifically, we evaluate the runtime on the TUD-L, 3DMatch, KITTI, and MVP-RG datasets using DCP, GeoTransformer, and LoGDesc as backbones, respectively. All models are executed on the same NVIDIA RTX A5000 GPU, and the results summarized in~\cref{tab.time} show that our method has approximately the same cost time as the Sinkhorn algorithm.

\begin{table}[!h]\scriptsize
    % \footnotesize
    \caption{Time cost of \textit{iMatcher} \textit{vs} Sinkhorn on different datasets.}
    \label{tab.time}
    \centering
    \begin{tabularx}{\linewidth}{X|X|X|X|X}
        \hline
        
        \hline
        
        \hline

        \bfseries Method & \multicolumn{4}{c}{ Mean inference time (seconds)} \\
        \cline{2-5}
        & \bfseries TUD-L & \bfseries 3DMatch  & \bfseries KITTI  & \bfseries MVP-RG \\
        \hline
        Sinkhorn & \textbf{0.15 }& \textbf{0.37} & 0.39  & \textbf{1.76} \\
        \hline
        \textbf{Ours} & 0.19 & 0.40 & \textbf{0.37}  & 1.79 \\
        \hline
        
        \hline
        
        \hline

    \end{tabularx}
\end{table}

% --------------------
\subsection{Ablation Study} \label{sec.ablation}
\textbf{KITTI.}  In this experiment, we investigate the impact of each \textit{iMatcher} component: Repositioning Step (\textbf{Rep}), global consistency encoding (\textbf{G.C}), bilateral matching features (\textbf{Bi. Match}) and local graph convolution module (\textbf{GCNN}). For the models without the repositioning step, the bidirectional source-to-target and target-to-source matching are performed by identifying the largest column-wise and row-wise values in the score matrix, respectively. The results reported in~\cref{imatcher_tab_ablation} show that at least four of the five used metrics are incrementally improved with the successive addition of each module. Notably, the final gap between the complete \textit{iMatcher} \textit{(e)} and vanilla version \textit{(a)} is $7.1\%$ in IR, while the RRE is reduced by $26\%$ and RTE by $16\%$.

\begin{table}[!h]\scriptsize
% \scriptsize
\caption{{Performance on KITTI. All methods get an FMR  of 99.8.} } 

% \centering
\resizebox{\columnwidth}{!}{%
\begin{tabularx}{\linewidth}{l l l l l | X | X | X | X | X}
\hline

\hline

\hline

& \textbf{GCNN}& \textbf{Bi. Match} &  \textbf{G.C} &  \textbf{Rep} &  IR & OV & RR & RRE & RTE \\

% \textbf{Lightglue }& \textbf{Sibkhirn} &  \textbf{G.C} &  \textbf{Rep} &  IR & OV & RR & RRE & RTE \\

% \textbf{Lightglue }& \textbf{NN Match} &  \textbf{G.C} &  \textbf{Rep} &  IR & OV & RR & RRE & RTE \\

\hline
% Sinkhorn (GeoT.) & \textbf{99.8} & 68.4 & 93.7 & \textbf{99.8} & 0.230 & 0.062 \\ 
% Sinkhorn & 75.7  &  -  & \textbf{99.8}  &  0.23 & 6.2\\
\textit{(a)} &\ding{55} & \ding{55} & \ding{55} &  \ding{55} & 88.0 & 93.8 & \textbf{99.8} & 0.27 & 5.7 \\ 
\textit{(b)} &\ding{51} &  \ding{55} & \ding{55} &  \ding{55}& 88.5 & 93.9 & 99.6 & 0.26 & 5.4 \\ 
\textit{(c) }& \ding{51} & \ding{51} & \ding{55} & \ding{55} & 92.7 & 97.1 & \textbf{99.8} & 0.21 & 5.0 \\ 
\textit{(d)} & \ding{51}  & \ding{51}  & \ding{51} & \ding{55} & 93.4 & 97.4 & \textbf{99.8} & 0.21 & \textbf{4.8} \\ 
\textit{(e)} & \ding{51}  & \ding{51} & \ding{51}  & \ding{51}  & \textbf{95.1} & \textbf{98.2} & \textbf{99.8} & \textbf{0.20} & \textbf{4.8} \\ 

\hline

\hline

\hline

\end{tabularx}%
}
\label{imatcher_tab_ablation}
\end{table}

First, we propose studying the variations in the \textit{iMatcher} performance under various numbers of neighbors used to build the local graphs, \textit{i.e.}, 8, 16, 32, and 40. As shown in~\cref{tab.knn}, using $32$ nearest neighbors yields the best results in both inlier ratio and pose estimation metrics (RRE and RTE).
\begin{table}[!h]\scriptsize
\centering
% \footnotesize
\caption{Performance on KITTI using different numbers of neighbors in the local graphs.}
\label{tab.knn}
\begin{tabularx}{\linewidth}{l | X | X | X | X}

\hline

\hline

\hline

{KNN} & {IR ($\%$) $\uparrow$} & {RR ($\%$) $\uparrow$} & {RRE (deg) $\downarrow$} & {RTE (cm) $\downarrow$} \\
\hline
 8 & 94.8 & \textbf{ 99.8}& \textbf{0.20} & 5.0 \\
% 12 & 95.1 & \textbf{ 99.8} & \textbf{0.20} & 5.0 \\
16 & 95.1 & \textbf{ 99.8}& \textbf{0.20} & 4.9 \\
32 & \textbf{95.1} &\textbf{ 99.8} & \textbf{0.20} & \textbf{4.8} \\
40 & 94.9 &  \textbf{ 99.8} &  \textbf{0.20} & \textbf{4.8} \\

\hline

\hline

\hline

\end{tabularx}

\end{table}

\noindent
\textbf{KITTI-360.} % pttre object-centric si y a tudl/MVP
We extend the previous ablation study by applying the same modifications to sequences 9 and 10 of KITTI-360 (1309 evaluation examples).~\cref{imatcher_tab_ablation360} shows that each \textit{iMatcher} compound significantly improves IR, OV, RRE, and RTE, except for the repositioning step, which slightly deteriorates RTE compared to KITTI while improving the other metrics. Notably, 100\% of the registrations are successful with the addition of the first compound \textbf{GCNN}, and this result holds for all compounds.
\begin{table}[!h]\scriptsize
% \scriptsize
\caption{{Performance on the KITTI-360 dataset.} } 

% \centering
\resizebox{\columnwidth}{!}{%
\begin{tabularx}{\linewidth}{l l l l l | X | X | X | X | X}
\hline

\hline

\hline

& \textbf{GCNN}& \textbf{Bi. Match} &  \textbf{G.C} &  \textbf{Rep} &  IR & OV & RR & RRE & RTE \\

\hline
% Sinkhorn (GeoT.) & \textbf{99.8} & 68.4 & 93.7 & \textbf{99.8} & 0.230 & 0.062 \\ 
% Sinkhorn & 75.7  &  -  & \textbf{99.8}  &  0.23 & 6.2\\
\textit{(a)} & \ding{55} & \ding{55} & \ding{55} &  \ding{55}   & 87.4 & 93.7  & 99.9           & 0.285    & 9.3   \\ 
\textit{(b)} & \ding{51} &  \ding{55} & \ding{55} &  \ding{55}  & 88.5 &  94.6 & \textbf{100}   &  0.271   &   8.6 \\ 
% \ding{51} & \ding{51} & \ding{55} & \ding{55}    & 91.6 &  97.2  & \textbf{100}  &  0.245   &  8.4  \\ 
\textit{(c)} & \ding{51}  & \ding{51}  & \ding{55} & \ding{55}  &  90.4 &  96.7 &  \textbf{100} &  0.249    &  8.3  \\
\textit{(d)} & \ding{51}  & \ding{51}  & \ding{51} & \ding{55}  &  93.4 &  97.4 &  \textbf{100} &  0.245    &  \textbf{8.1 } \\
% [2025-02-19 13:07:24] [CRIT]   Fine Matching, FMR: 1.0000, IR: 0.934, IR_0.3: 0.358, IR_0.1: 0.036, num_Corr: 2786.088, OV: 0.974, std: 0.000
% [2025-02-19 13:07:24] [CRIT]   Registration, RR: 1.0000, RRE: 0.245, RTE: 0.084, Rx: 0.178, Ry: 0.081, Rz: 0.356
\textit{(e)} & \ding{51}  & \ding{51} & \ding{51}  & \ding{51}  & \textbf{94.5} & \textbf{98.2} &\textbf{100}   & \textbf{0.242} & {8.2} \\ 

\hline

\hline

\hline

\end{tabularx}%
}
\label{imatcher_tab_ablation360} 
\end{table}

\begin{figure}[!h]
\centerline{\includegraphics[width=1\columnwidth]{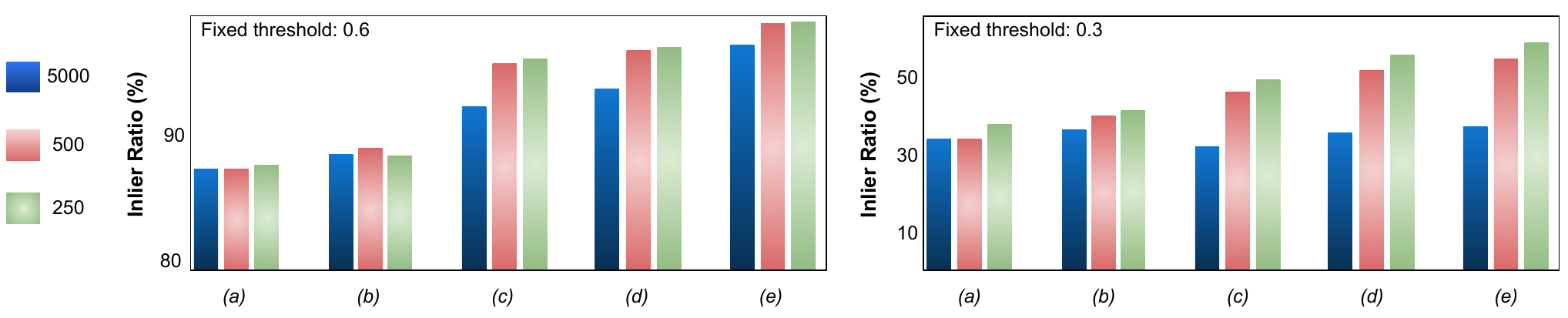}}

\caption{Inlier ratio evolution for the different ablated pipelines on KITTI-360, with $\eta =0.6$ (top) and $\eta =0.3$ (bottom)  }
\label{fig.inlier_kitti360}
\end{figure}

~\cref{fig.inlier_kitti360} depicts the variations in inlier ratio (IR) under two different thresholds, \textit{i.e.}, 0.6 and 0.3, using various numbers of putative correspondences: 250, 500, and 5000 for all the models described in~\cref{imatcher_tab_ablation360}.

\section{Discussion}\label{matching_sec_discussion}
% -------------
%
The \textit{iMatcher} pipeline, proposed in the previous sections, offers a fully differentiable and interpretable framework for feature matching in rigid registration tasks. Extensive experiments across diverse configurations, including indoor and outdoor scenes as well as object-centric datasets, demonstrate its versatility and robustness. Notably, \textit{iMatcher} performs well under challenging conditions, such as low-overlap scenarios (\textit{e.g.}, down to 30\% overlap on the 3DMatch benchmark). Aside from the local neighborhood size, the method requires no additional parameter tuning, as its learning-based design effectively regulates the sensitivity of the global geometric compatibility matrix. The inlier ratio has been improved in almost all tested settings, and \textit{iMatcher} achieved state-of-the-art performance on this metric by a large margin in most cases. 

The ablation studies presented in~\cref{sec.ablation} demonstrate the substantial contribution of each proposed component. In particular, we observe a significant performance gap between the initial matrix $\hat{S}$ (which excludes score learning, the repositioning step, and bilateral matching) and the final refined matrix $S$. Across all evaluation metrics, $S$ consistently outperforms $\hat{S}$. On the KITTI dataset, for instance, we observe improvements of +6.6\% in inlier ratio (IR), +23\% in relative rotation error (RRE), and +11\% in relative translation error (RTE).

One structural limitation of \textit{iMatcher} lies in the computation of the confidence value in Equation~\cref{imatcher_alpha_mean}. For extremely low overlapping cases and in the presence of enormous outliers in the initial putative correspondences, the weight $\alpha$ obtained using an average pooling over all the correspondences may be overwhelmed and lead to less sparsity between the outliers and inliers' confidence values.  One other limitation can be noticed. In some cases, high inlier ratios do not necessarily translate into improved registration performance. For instance, in low-overlap scenarios such as those in the 3DMatch dataset, the registration performance appears to saturate once a certain inlier ratio threshold is reached. This phenomenon is evident in~\cref{tab.3dmatch}, where a relative improvement of over 15\% in inlier ratio results in only a marginal increase in registration recall (\textit{i.e.}, +0.2\%).  To further investigate these two effects, we extend our experimental study by evaluating the method on the more challenging 3DLoMatch benchmark. This subset of 3DMatch comprises only point cloud pairs with less than 10\% overlap, offering a more stringent test of registration robustness under extreme conditions. Following~\cite{geotransformer,yu2021cofinet}, the same model trained on the complete 3DMatch is used in this case, and the results are reported in~\cref{iMatcher_tab_3dlomatch}.

\begin{table}[!h]
\scriptsize  % Makes the table more compact
\caption{Performance on the 3DLoMatch dataset.}\label{iMatcher_tab_3dlomatch}
\centering
\begin{tabularx}{\linewidth}{X|X|X|X}

    \hline 

    \hline 
    
    \hline
    
\hline
% \textbf{Model}  & \textbf{IR(\%)} & \textbf{RR (RANSAC) (\%)} & \textbf{RR (SVD) (\%)} \\
\textbf{Model} & \textbf{IR(\%)} & \makecell{\textbf{RR}\textbf{(\%)}\\\textbf{RANSAC}} & \makecell{\textbf{RR}\textbf{(\%)}\\\textbf{SVD}} \\

\hline
D3Feat~\cite{bai2020d3feat}  & 13.2 & 37.2 & 2.8 \\
% SpinNet~\cite{SpinNet}  & 20.5 & 59.8 & 2.5 \\
Predator~\cite{huang2021predator}  & 26.7 & 69.0 & 6.4 \\
CoFiNet~\cite{yu2021cofinet}  & 24.4 & 67.5 & 21.6 \\
GeoT. (Sink.)~\cite{geotransformer}  & \underline{44.4} & \textbf{73.9} &\underline{61.1} \\
GeoTr. (\textit{iMatcher})  & \textbf{64.4} & \underline{71.1} & \textbf{67.1} \\
\rowcolor{blue!10} \textcolor{blue}{\emph{{Improvement ($\%$)}}} &    \textit{\textcolor{blue}{+45.0}}  & \textit{\textcolor{black}{-3.8}}  & \textit{\textcolor{blue}{+9.8}}  \\
% \hline
    \hline 

    \hline 
    
    \hline
    
\end{tabularx}
\end{table}

% -------------------
\section{Conclusion}
% -------------------

In this paper, we introduced \textit{iMatcher}, a fully differentiable framework for robust and accurate point cloud feature matching. Our approach utilizes local graph convolutions to enhance point cloud features and initialize the score matrix, effectively capturing local structures and enabling a repositioning step of the source point cloud. To enhance inlier-outlier distinction, we introduce a point-wise match probability learning module that integrates global spatial consistency-aware feature encoding, guided by bidirectional source-target and target-source pairings. The final relaxed stochastic similarity mapping is then estimated by combining these components. Extensive evaluations on outdoor and indoor scans, object-centric registration, and partial-to-partial matching benchmarks demonstrate that our method significantly outperforms state-of-the-art approaches, particularly by improving the predicted inlier ratio. For future work, we plan to incorporate a denoising diffusion process into our model to better handle highly noisy and low-overlap data. Additionally, we aim to explore its adaptability to non-rigid shape matching by leveraging local rigidity in deformations as an alternative to the proposed global consistency-aware feature embedding.

\section{Acknowledgements}
This work was supported by the French ANR program MARSurg (ANR-21-CE19-0026).

\bibliography{main}
\end{document}